\documentclass[conference]{IEEEtran}
\IEEEoverridecommandlockouts

\usepackage{cite}
\usepackage{amsmath,amssymb,amsfonts}
\usepackage{algorithmic}
\usepackage{graphicx}
\usepackage{textcomp}
\usepackage{xcolor}

\usepackage{todonotes}
\usepackage{subfig}
\usepackage{array}
\usepackage{tikz}

\newcommand\cross[2]{%
  \multicolumn{1}{p{#1}}{\hskip-\tabcolsep
  $\vcenter{
  \begin{tikzpicture}[baseline=0,anchor=south west]
  \path[use as bounding box] (0,0) rectangle (#1+2\tabcolsep,\baselineskip);
  \node[minimum width={#1+2\tabcolsep},minimum height=\baselineskip+\extrarowheight] (box) {};
  \draw (box.north west) -- (box.south east);
  \draw (box.north east) -- (box.south west);
  \node[anchor=center] at (box.center) {#2};
 \end{tikzpicture}
 }$\hskip-\tabcolsep}}

\usepackage{url}

\def\BibTeX{{\rm B\kern-.05em{\sc i\kern-.025em b}\kern-.08em
    T\kern-.1667em\lower.7ex\hbox{E}\kern-.125emX}}

\begin{document}

\title{Uncertainty-Preserving QBNNs: Multi-Level Quantization of SVI-Based Bayesian Neural Networks for Image Classification\\
\thanks{\textsuperscript{*}These authors contributed equally to this work.}
}

\author{\IEEEauthorblockN{Hendrik Borras*, Yong Wu*, Bernhard Klein, Holger Fröning}
	\IEEEauthorblockA{\textit{Hardware and Artificial Intelligence (HAWAII) Lab, Heidelberg University, Germany} \\
		\{hendrik.borras, bernhard.klein, holger.froening\}@ziti.uni-heidelberg.de, yongwu\_cs@outlook.com}
}

\maketitle

\begin{abstract}
Bayesian Neural Networks (BNNs) provide principled uncertainty quantification but suffer from substantial computational and memory overhead compared to deterministic networks.
While quantization techniques have successfully reduced resource requirements in standard deep learning models, their application to probabilistic models remains largely unexplored.
We introduce a systematic multi-level quantization framework for Stochastic Variational Inference based BNNs that distinguishes between three quantization strategies: Variational Parameter Quantization (VPQ), Sampled Parameter Quantization (SPQ), and Joint Quantization (JQ).
Our logarithmic quantization for variance parameters, and specialized activation functions to preserve the distributional structure are essential for calibrated uncertainty estimation.
Through comprehensive experiments on Dirty-MNIST, we demonstrate that BNNs can be quantized down to 4-bit precision while maintaining both classification accuracy and uncertainty disentanglement.
At 4 bits, Joint Quantization achieves up to 8$\times$ memory reduction compared to floating-point implementations with minimal degradation in epistemic and aleatoric uncertainty estimation.
These results enable deployment of BNNs on resource-constrained edge devices and provide design guidelines for future analog "Bayesian Machines" operating at inherently low precision.

\end{abstract}


\section{Introduction}

Deep Neural Networks (DNNs) have become the dominant modeling paradigm across vision, language, and scientific domains, yet they still lack a principled mechanism to quantify uncertainty in their predictions.
Standard architectures produce point estimates---often interpreted as probabilities through functions such as softmax—even though these outputs do not reflect true probabilities~\cite{guo2017calibrationmodernneuralnetworks}.
Consequently, neural networks may remain overconfident when confronted with inputs that lie outside their domain of competence (out-of-domain data).

Bayesian Neural Networks (BNNs)~\cite{mackay1992practical,neal1996bayesian} address this limitation by placing probability distributions over model parameters, allowing them to quantify uncertainty in their predictions and to differentiate between uncertainty inherent in the data (aleatoric uncertainty) and uncertainty that reflects the model's lack of knowledge or capacity (epistemic uncertainty)~\cite{kendall2017uncertaintiesneedbayesiandeep,Jospin_2022}.
This probabilistic formulation provides a mathematically grounded way for models to express limited knowledge, which is crucial under distribution shift, scarce training data, or safety-critical decision-making~\cite{Jospin_2022}.
Among approximate inference techniques for BNNs, Stochastic Variational Inference (SVI) has emerged as a practical and scalable approach due to its compatibility with modern deep learning frameworks~\cite{Blei_2017,blundell2015weight}.

Despite these advantages, BNNs suffer from substantial computational and memory overhead.
The need to maintain and update distributions over parameters---for SVI typically represented via floating-point distributional parameters such as means and variances of Gaussians---results in significantly higher memory and compute requirements than in standard deterministic networks.
This mismatch limits deployment on resource-constrained hardware, where power, throughput, and memory bandwidth are critical constraints.

Neural network quantization provides a powerful way to reduce model size and compute cost by lowering numerical precision~\cite{JMLR:v25:18-566}.
However, while quantization is well understood for deterministic models, applying it directly to probabilistic models is non-trivial: quantization may distort the distributional structure essential for calibrated uncertainty estimates.
Recent works have begun to explore low-precision Bayesian neural networks, including post-training quantization of variational BNNs~\cite{ferianc2021quantisation,subedar2021quantization,lin2023quantization}. 
However, these approaches treat quantization as a single, post-hoc compression step and do not analyze where and how quantization should be applied within the SVI pipeline to preserve predictive uncertainty. 
In particular, we are not aware of a multi-level, SVI-integrated quantization framework that studies bit-width vs. uncertainty fidelity (aleatoric and epistemic) across inputs, variational parameters, and stochastic samples.
Three fundamental gaps persist:
\begin{enumerate}
    \item No multi-level view of quantization in Bayesian inference: existing work treats quantization as a single operation, ignoring the different stochastic levels in SVI (inputs, distributions, samples).
    \item Lack of uncertainty-preserving precision reduction methods: there are no quantization approaches explicitly designed to preserve the statistical semantics of $\mu$, $\sigma$, and sampled weights in BNNs.
    \item No systematic evaluation of bit-width vs. uncertainty fidelity: prior studies mainly examine accuracy vs. bit-width, but not the impact on aleatoric and epistemic uncertainty, nor on uncertainty calibration.
\end{enumerate}
Addressing these gaps is essential to enable hardware-efficient Bayesian inference that maintains the main benefit of BNNs: reliable uncertainty estimation.

The objective of this work is to develop and experimentally validate a hardware-aware, multi-level quantization framework for SVI-based Bayesian Neural Networks that reduces computational cost while preserving accuracy and calibrated predictive uncertainty.
Concretely, our contributions are:
\begin{enumerate}
	\item \textbf{Multi-Level Quantization Framework for SVI-BNN Classifiers}: we introduce a systematic decomposition of quantization locations in SVI-based Bayesian image classifiers—covering inputs, sampled weights, and variational distribution parameters—and analyze how each level affects classification accuracy and predictive uncertainty (aleatoric and epistemic) on Dirty-MNIST~\cite{Mukhoti2022dirtyMNIST}.
    \item \textbf{Quantization-Aware SVI Optimization}: we design and empirically study quantization-robust SVI configurations, including specialized activation functions, as well as magnitude-based clipping and log-quantization strategies that preserve the variance structure of latent distributions under low precision.
    \item \textbf{Comprehensive Bit-Width Sensitivity Study}: we perform an extensive analysis of classification accuracy, and aleatoric/epistemic uncertainty decomposition across multiple bit-widths.
    This yields practical guidelines for selecting precision levels that balance efficiency and uncertainty fidelity.
\end{enumerate}

The ability to deploy Bayesian models on constrained hardware unlocks applications where both low latency and calibrated uncertainty are required, such as embedded medical devices, autonomous robots, and safety-critical control systems.
This work bridges the gap between reliable probabilistic modeling and low-precision hardware efficiency, and provides a first step toward principled design rules for quantized Bayesian inference.
The insights are directly relevant for future BNN accelerators (called "Bayesian Machines"), for example based on probabilistic photonic computing~\cite{chaotic_nature_2024,photonic_prob_ai_outlook_2025,TWI_pre_print_2025}, FPGA implementations, and mixed-signal/neuromorphic platforms that naturally operate at low bit-widths or with stochastic hardware primitives.

\section{Background}\label{sec:background}
\begin{figure}[htbp]
	\centering
	\includegraphics[width=0.7\linewidth]{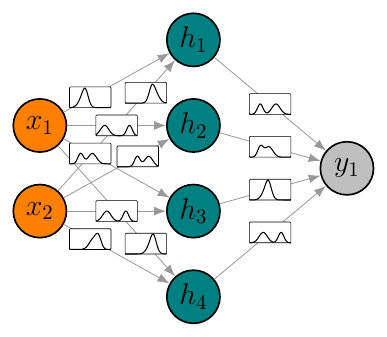}
	\caption{A Bayesian Neural Network can be viewed as a straightforward extension of a standard neural network, where the deterministic parameters are replaced by probability distributions.}
	\label{fig:background:bnn}
\end{figure}

Bayesian Neural Networks are widely regarded as a principled approach to uncertainty-aware learning \cite{Ghahramani2015, blundell2015weight}.
They aim to approximate the posterior distribution 
\( p(\omega | D) \propto p(D | \omega)\, p(\omega) \) over parameters \( \omega \).
Since exact posterior inference is intractable for modern neural architectures, a variety of approximate methods have been developed, including Variational Inference (VI)~\cite{Blei_2017,blundell2015weight}, Markov chain Monte Carlo (MCMC), Deep Ensembles~\cite{deepensembles}, and Monte Carlo dropout~\cite{gal2016dropout}.
In VI, one introduces a parametrized family \( q_\Theta(\omega) \) over the model parameters \( \omega \) (see also Figure~\ref{fig:background:bnn}), where \( \Theta \) denotes the variational parameters, and optimizes the evidence lower bound (ELBO) to minimize the Kullback--Leibler divergence between \( q_\Theta(\omega) \) and the true posterior \( p(\omega | \mathcal{D}) \).

In contrast to sampling-based methods such as Markov chain Monte Carlo, which are computationally prohibitive for large-scale data, VI casts posterior estimation as an optimization problem that can be solved efficiently using stochastic gradient updates \cite{hoffman2013stochastic}.
In practice, one often assumes independent Gaussian distributions parametrized by mean $\mu$ and variance $\sigma^2$ for \( q_\Theta(\omega) \), following the mean-field assumption~\cite{Jospin_2022}.

\subsection{BNN Quality Metrics}

While general uncertainty assessments are useful, BNNs often benefit from a more fine-grained view, separating uncertainty into \emph{aleatoric} and \emph{epistemic} components.
Aleatoric uncertainty refers to variability arising from noise in the underlying data-generating process.
Because this uncertainty is inherent to the observations, it cannot be reduced even when additional data are collected.
Epistemic uncertainty, in contrast, reflects the modeling error and typically decreases as more data become available.
It captures aspects of model capacity and learned knowledge, and is commonly assessed by evaluating the model’s ability to detect out-of-domain (OOD) data.

This decomposition of aleatoric and epistemic uncertainty separates total uncertainty into two distinct scores:
\begin{itemize}
\item Softmax Entropy ($\mathbb{E}_{p(w|\mathcal{D})}[H[y|x, w]]$) represents the aleatoric uncertainty component.
It quantifies the average predictive ambiguity conditioned on a specific model, capturing the inherent noise or class overlap present in the input.
\item Mutual Information ($I[y, w|x, \mathcal{D}]$) quantifies the epistemic uncertainty.
It quantifies how much the predictive distribution varies across posterior weight samples, capturing uncertainty arising from insufficient data or imperfectly learned models.
\end{itemize}

In practice, these uncertainty components can be estimated using Monte Carlo sampling from the posterior distribution, i.e., executing multiple forward passes of the Bayesian Neural Network to obtain multiple predictions (samples) $p(y=c|x,w_n)$.
For brevity, let $p$ denote $p(y=c|x,w_n)$.
With $N$ weight samples $\{w_n\}_{n=1}^{N}$ from the approximate posterior $q(w)$, the \emph{Softmax Entropy} (aleatoric) is computed as:
{%
\begin{equation}
	\mathbb{E}_{p(w|\mathcal{D})}[H[y|x, w]] \approx -\frac{1}{N}\sum_{n=1}^{N}\sum_{c=1}^{C} p \cdot \log p
\end{equation}
}
This expression calculates the average entropy of the per-sample softmax outputs.
The total uncertainty is estimated by first averaging the softmax probabilities across samples to obtain the predictive distribution, then computing its entropy:
{%
\begin{equation}
	H(y|x, \mathcal{D}) \approx -\sum_{c=1}^{C} \left( \Bigl( \frac{1}{N}\sum_{n=1}^{N}p\Bigr) \cdot \log \Bigl(\frac{1}{N}\sum_{n=1}^{N}p \Bigr) \right)
\end{equation}
}
Finally, the \emph{Mutual Information} (epistemic) is derived by subtracting the aleatoric component from the total uncertainty:
\begin{equation}
	I[y, w|x, \mathcal{D}] = H(y|x, \mathcal{D}) - \mathbb{E}_{p(w|\mathcal{D})}[H[y|x, w]]
\end{equation}

\subsection{BNN Evaluation Dataset}
\label{sec:dirtymnist}
\begin{figure}[htbp]
	\centering
	\includegraphics[width=0.9\linewidth]{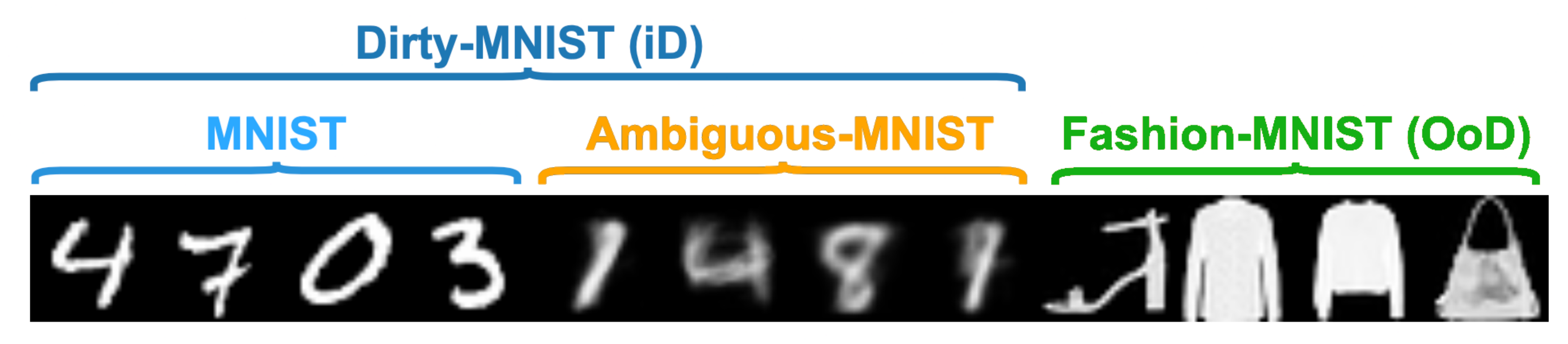}
	\caption{Examples from the Dirty-MNIST dataset as used for assessment of in-domain prediction quality (MNIST), aleatoric uncertainty (Ambiguous-MNIST) and epistemic uncertainty (Fashion-MNIST), respectively.}
	\label{fig:meth:dmnist}
\end{figure}

The primary community standard for evaluating a BNN for epistemic and aleatoric uncertainty is Dirty-MNIST\cite{Mukhoti2022dirtyMNIST},
where the whole dataset is composed of a set of three sub-datasets, as illustrated in Figure \ref{fig:meth:dmnist}.
The first set being standard MNIST \cite{726791} and Fashion-MNIST \cite{Fashion-MNIST},
where Fashion-MNIST is used as the OOD test case.
Additionally, Dirty-MNIST introduces Ambiguous-MNIST~\cite{Mukhoti2022dirtyMNIST}, which is constructed such that each image resembles two digits and thus classes simultaneously.
Meaning, that it can be used to evaluate aleatoric uncertainty, as the distinction between the two classes is irreducible.

Training is then done on the training sets of both MNIST and Ambiguous-MNIST, while evaluation is done using the test splits of all three datasets.
The resulting prediction quality is judged using multiple metrics:
\begin{itemize}
	\item In-domain prediction quality is evaluated using accuracy evaluation on MNIST.
	\item Aleatoric uncertainty estimation quality is evaluated using the Area Under the Receiver Operating Characteristic Curve (AUROC) of MNIST against Ambiguous-MNIST.
	\item Epistemic uncertainty estimation quality is similarly evaluated using the AUROC of MNIST and Ambiguous-MNIST against Fashion-MNIST.
	\item To investigate the disentanglement of aleatoric and epistemic uncertainty, we visually investigate how the three datasets separate in a scatter plot of Mutual Information over Softmax Entropy.
\end{itemize}
Using these metrics and factors one is able to concisely investigate the prediction quality not only of the standard quality metric for classification (accuracy), but also evaluate how well uncertainty estimation and disentanglement work.
In a prototypical scatter plot (e.g. Figure~\ref{fig:challenge:act_fun_compare}), all samples from Ambiguous-MNIST should cluster on the x-axis (increased Softmax Entropy score but Mutual Information close to zero), while the samples from Fashion-MNIST should cluster around the y-axis (increased Mutual Information score but Softmax Entropy close to zero). 
Samples from MNIST should have low scores for both Mutual Information and Softmax Entropy, thus should cluster in the bottom left of the figure.

\section{Related Work}
Although there is a vast body of work on quantization for standard Deep Neural Networks, BNNs have received surprisingly little dedicated attention.
While one might expect that prediction accuracy under quantization behaves similarly for BNNs and deterministic DNNs, the crucial open question is how quantization impacts \emph{uncertainty estimates}---the very core reason for using BNNs in the first place.

Quantization for standard Deep Neural Networks has been extensively studied in the context of resource-efficient inference on embedded and general-purpose hardware.
Surveys and system studies such as Roth et al.~\cite{JMLR:v25:18-566} and Gholami et al.~\cite{Gholami2021} provide overviews of compression techniques---especially Post-Training Quantization (PTQ), Quantization-Aware Training (QAT), and mixed-precision schemes---and document that 4--8\,bit inference is often achievable with minimal accuracy loss on standard benchmarks.
Foundational work by Jacob et al.~\cite{Jacob2018} introduced the now-standard \emph{integer-arithmetic-only 8-bit quantization} with scale and zero-point, forming the basis of many mobile and edge deployments, while Banner et al.~\cite{Banner2019} demonstrated that, with careful activation clipping and bias correction, even \emph{4-bit PTQ} can be practical for convolutional networks without retraining.
Beyond uniform bit-widths, methods such as HAQ~\cite{Wang2019} or Galen~\cite{krieger2023galen} treat quantization as a \emph{hardware-aware optimization problem}, automatically selecting layer-wise bit-widths under latency, energy, or model-size constraints for specific accelerators.
System-focused frameworks like DeepChip~\cite{DBLP:conf/pkdd/SchindlerZPF18} integrate low-precision arithmetic with sparsity and optimized kernels to achieve multi-$\times$ speedups and substantial memory savings on constrained hardware.
Together, these works underscore the maturity and effectiveness of low-bit quantization for deterministic DNNs.

Ferianc et al.~\cite{ferianc2021quantisation} conduct a broad empirical investigation into how low-precision quantization affects both predictive accuracy and uncertainty quality in BNNs.
They evaluate several Bayesian inference schemes---including Bayes-by-Backprop, MC Dropout, and SGHMC---under uniform quantization of weights and activations down to sub-8-bit precision across a range of datasets and architectures.
Their main finding is that BNNs are surprisingly robust to standard low-bit quantization: predictive performance, calibration, and uncertainty measures remain largely stable even at aggressive bit-width reductions.
However, their study focuses on generic uniform PTQ and does not analyze how quantization interacts with the internal structure of variational inference or task-specific pipelines, leaving open questions about where quantization most strongly affects uncertainty propagation in SVI-based Bayesian classifiers.

Subedar et al.~\cite{subedar2021quantization} present one of the first empirical investigations into the quantization of SVI-BNNs, focusing on PTQ of pretrained mean-field variational models on MNIST and CIFAR-10.
Their method quantizes the learned posterior parameters by applying INT8 PTQ to the mean and a uniform low-bit quantizer to the standard deviation, while also quantizing sampled noise for Monte Carlo inference.
Their study demonstrates that even aggressive variance quantization (down to 1\,bit) preserves predictive accuracy and calibration, and does not substantially degrade uncertainty quality, even under dataset shift.
However, their approach is limited to a single-level, post-hoc quantization of the trained posterior and does not modify the SVI training process, the stochastic sampling pathway, or the input representation.

Lin et al.~\cite{lin2023quantization} extend the line of work on quantized Bayesian deep learning by proposing a PTQ workflow for BNNs built on Bayesian-Torch and targeting INT8 inference on 4th Gen Intel Xeon (Sapphire Rapids).
Their framework mirrors standard \textit{PyTorch} static PTQ: observers are inserted, calibration is performed on representative data, and full-precision Bayesian models (e.g., variational ResNet-50 on ImageNet) are converted into quantized BNNs with INT8 weights and activations that exploit Intel AMX instructions.
They then characterize low-precision BNN workloads at system level, reporting up to $6.9\times$ inference throughput speedup and $4\times$ memory reduction over FP32 BNNs while preserving top-1 accuracy and uncertainty calibration on ImageNet.
Beyond ImageNet, they evaluate a medical histology classifier and OOD detection (Camelyon17-WILDS), showing that robustness to data drift and the quality of predictive uncertainty are essentially unaffected by INT8 quantization.

Taken together, the studies by Ferianc et al., Subedar et al., and Lin et al.\ demonstrate that Post-Training Quantization of pretrained Bayesian models---including INT8 quantization of weights and activations and even sub-8-bit quantization of $\sigma$---can preserve accuracy and aggregate uncertainty metrics on standard vision benchmarks.
These methods, however, all treat quantization as a single-step, post-hoc compression procedure applied to a fixed Bayesian model, and primarily consider uniform or INT8 schemes.
In contrast, our work performs a multi-level, intra-SVI quantization analysis for Bayesian image classification on Dirty-MNIST.
Rather than quantizing $\mu$ and $\sigma$ alone, we explicitly quantize distribution parameters, and stochastic samples, while additionally introduce quantization-aware SVI adjustments to examine how different quantization locations and bit-widths affect predictive uncertainty (aleatoric and epistemic).
Furthermore, unlike Lin et al., we do not target one specific INT8 hardware pipeline but instead develop hardware-aware yet hardware-agnostic guidelines for sub-8-bit Bayesian classification.

\section{Multi-level quantization of Bayesian Neural Networks}\label{sec:multi-level}

Compared to the quantization of deterministic Deep Neural Networks, 
Bayesian Neural Networks require more than just straightforward quantization of weights and activations.
For these one must distinguish between values which are sampled from probability distributions and those which are not.
Since probability distributions in SVI are always parametrizable distributions, there are two points at which quantization can occur, resulting in three types of possible quantization strategies:
\begin{enumerate}
	\item \textbf{Variational Parameter Quantization (\textit{VPQ}):} Quantizing the parameters of the variational distribution, such as the mean and variance in a Gaussian, reduces the memory footprint of the model.
	The sampled values themselves, however, remain in floating-point format, meaning that the resulting computations are not inherently more efficient than those of standard variational inference.
	\item \textbf{Sampled Parameter Quantization (\textit{SPQ}):} Instead of quantizing the variational parameters, this approach quantizes the samples drawn from the variational distribution, enabling integer computations when activations are quantized as well.
	\item \textbf{Joint Quantization (\textit{JQ}):} This approach combines both \textit{VPQ} and \textit{SPQ} to simultaneously reduce memory consumption and computational cost.
	It therefore constitutes the primary method investigated in this work.
\end{enumerate}
For the standard implementation of BNNs, in which weights are probabilistic this means the following:
activations can be quantized just as in deterministic DNNs, but weights need to be taken into special consideration, given the points above.

For the implementation of the investigated BNN we use \textit{Pyro} \cite{bingham2018pyrodeepuniversalprobabilistic}, a probabilistic programming language (PPL), based on the widely-used deep learning framework \textit{PyTorch} \cite{paszke2019pytorchimperativestylehighperformance}.
\textit{Pyro} splits an SVI model into two parts:
\begin{itemize}
	\item \textbf{Model:} it contains the main computational graph, of the probabilistic model, for BNNs these are layers like fully-connected layers or activation functions.
	It additionally specifies in which places probabilistic distributions are sampled.
	For each distribution, a prior in the sense of Bayes' theorem needs to be given.
	\item \textbf{Guide:} it represents the variational distributions of SVI, i.e. the approximated Bayesian posterior.
	This means that the guide contains information about the structure of all probability distributions, whose parameters are learned during training.
\end{itemize}
During training this means that special attention needs to be paid to the guide in particular, as it contains the primary parameters of interest.
While some PPLs, such as \textit{Bayesian-Torch} and \textit{TensorFlow Probability}, do not do this explicit model/guide splitting. 
All PPLs implement a representation of the Bayesian prior and posterior, meaning that conclusions drawn from our \textit{Pyro} implementation apply similarly.

\begin{figure}[htbp]
	\centering
	\includegraphics[width=0.9\linewidth]{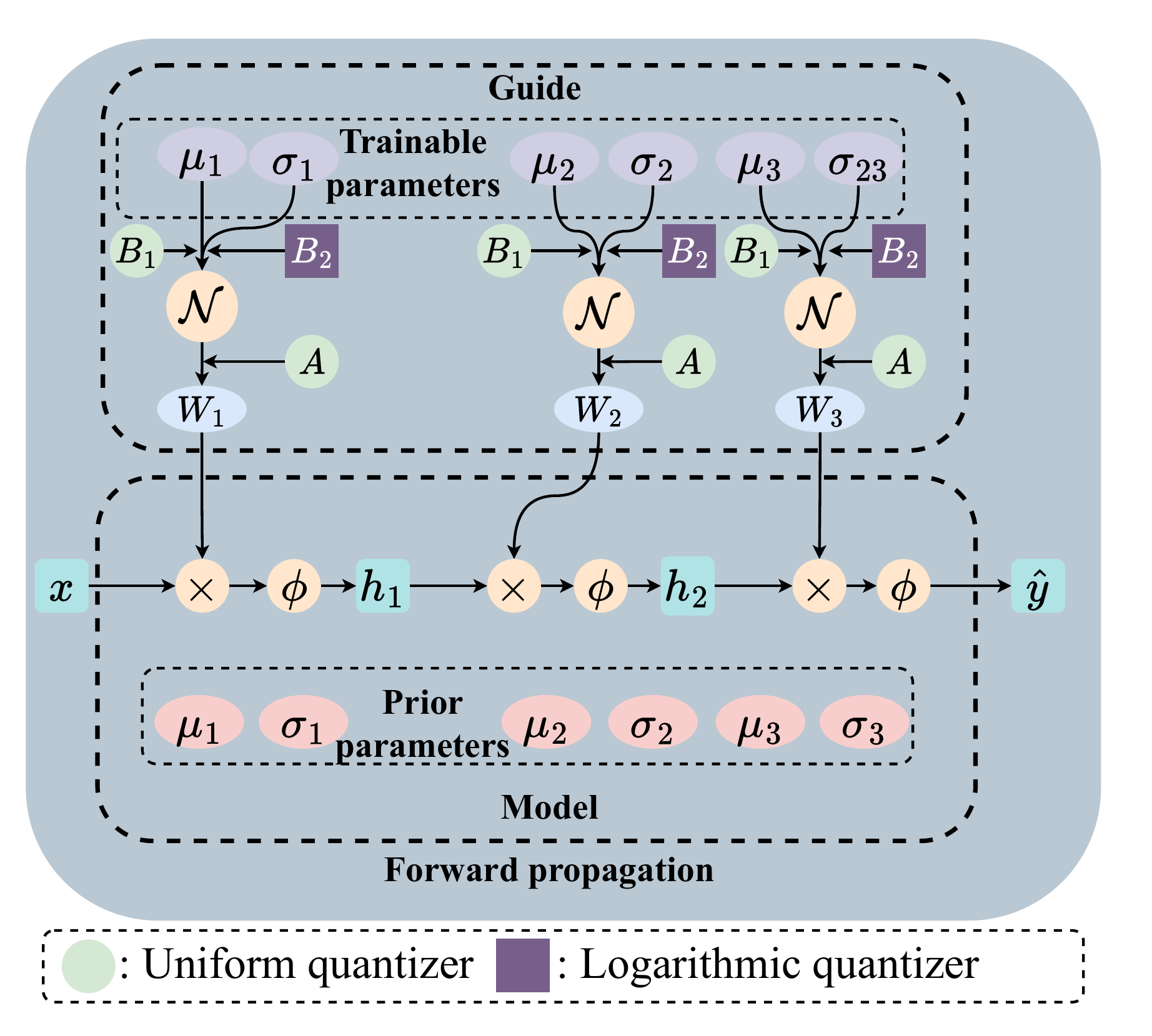}
	\caption{\textit{JQ} as implemented in \textit{Pyro}: uniform quantization is applied in general, while logarithmic quantization is applied specifically to standard deviation parameters.
	This approach allows for the more precise representation of a wide range of standard deviation values.}
	\label{fig:meth:quant}
\end{figure}

Figure \ref{fig:meth:quant} shows the quantization process used in this work for both the guide and model.
Notably we do not quantize activations, as we are primarily interested in the extension of quantization to BNNs, less the general interaction with DNN components.
As all parameters of the model live in the guide, the quantization is accordingly implemented there.
Each variational distribution is then equipped with three distinct quantizers:
\begin{itemize}
	\item Uniform quantizer $A$ (\textit{SPQ}): Quantizes the results of the sampling process 
	\item Quantizers $B$ (\textit{VPQ}): One for each parameter of the given variational distribution, here a Gaussian
	\begin{itemize}
		\item Uniform quantizer $B_1$: quantizes the mean parameter of the Gaussian distribution
		\item Logarithmic quantizer $B_2$: quantizes the variance parameter of the Gaussian distribution
	\end{itemize}
\end{itemize}
All quantizers are implemented as straight-through-gradient estimators, allowing for high-precision quantization-aware training.
From an implementation standpoint all uniform quantizers were implemented using the quantization library \textit{Brevitas} \cite{brevitas},
while the logarithmic quantizers are a custom implementation using \textit{PyTorch}.

\section{Challenges of quantized Bayesian Neural Networks}\label{sec:challenge}

As already touched upon in section \ref{sec:multi-level}: Quantization approaches in BNNs differ significantly from deterministic DNNs,
due to the introduction of distributions over weights instead of point estimates.
To further complicate things, BNNs are even in their non quantized form very sensitive to hyper-parameters.

\subsection{Activation functions for BNNs}

\begin{figure*}[tbh!]
    \centering
    \subfloat[ReLU]{
        \includegraphics[width=0.34\textwidth]{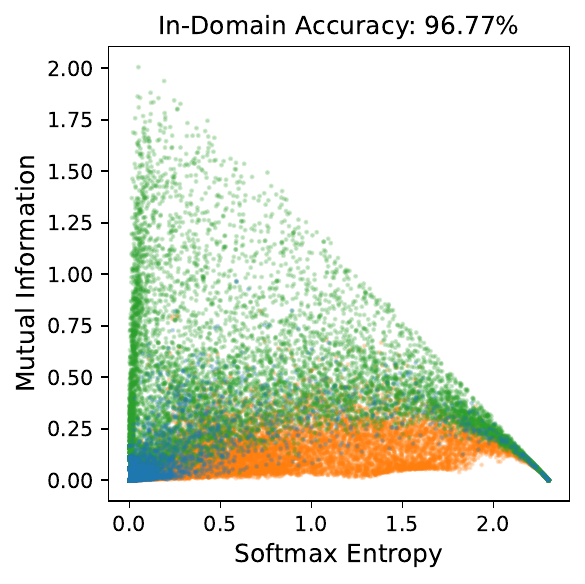}
        \label{fig:challenge:relu}
    }\hspace{3.em}
    \subfloat[SoftPlus]{
        \includegraphics[width=0.34\textwidth]{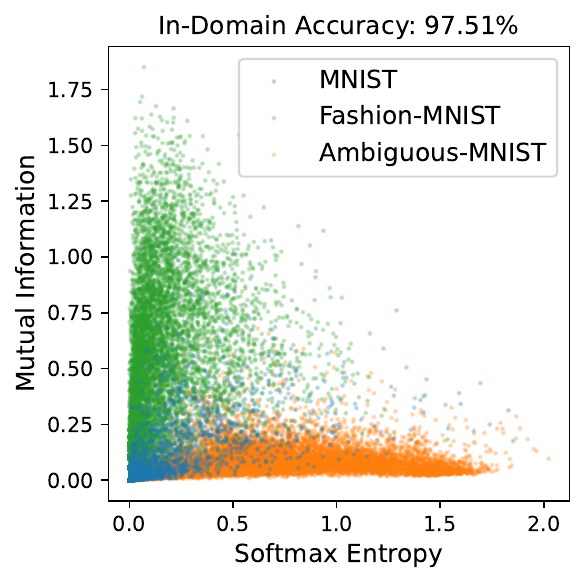}
        \label{fig:challenge:softplus}
    }
    \caption{Scatter plots comparing the relationship between Softmax Entropy and Mutual Information for Dirty-MNIST using different activation functions under 7-bit \textit{SPQ} quantization.
    SoftPlus (b) demonstrates superior uncertainty calibration compared to ReLU (a), showing more distinct clustering patterns for each dataset variant and maintaining clearer separation between different levels of aleatoric and epistemic uncertainty.}
    \label{fig:challenge:act_fun_compare}
\end{figure*}

In particular activation function choice is a significant concern for BNNs \cite{tempczyk2022simpletrickfixbayesian,klein2025phd}.
This is in stark contrast to deterministic networks, where especially for tasks such as classification only small changes, if any, have been made to the well-established baseline of the ReLU activation function.
To highlight this issue, Figure~\ref{fig:challenge:act_fun_compare} shows the same BNN from section \ref{sec:exp} with different activation functions,
while being otherwise quantized to seven bit, using \textit{SPQ} quantization.
At this level of quantization a good separation of aleatoric and epistemic uncertainty is still expected,
meaning that the three datasets should cluster into distinct regions of the scatter plot (also see Section~\ref{sec:dirtymnist}): MNIST on the bottom left, Fashion-MNIST on the top left and Ambiguous-MNIST on the bottom right.
While this works as expected for the SoftPlus activation function (Fig. \ref{fig:challenge:softplus}), the standard ReLU shows strong deficiencies in Fig. \ref{fig:challenge:relu}.
Notably, ReLU can still separate MNIST from the other datasets, but Fashion-MNIST and Ambiguous-MNIST become strongly entangled with each other.
Resulting in substandard uncertainty separation.
An additional benefit of SoftPlus is that the in-domain accuracy also slightly increases.

\subsection{Quantization of probability distributions}

\begin{figure}[htbp]
	\centering
	\includegraphics[width=0.9\linewidth]{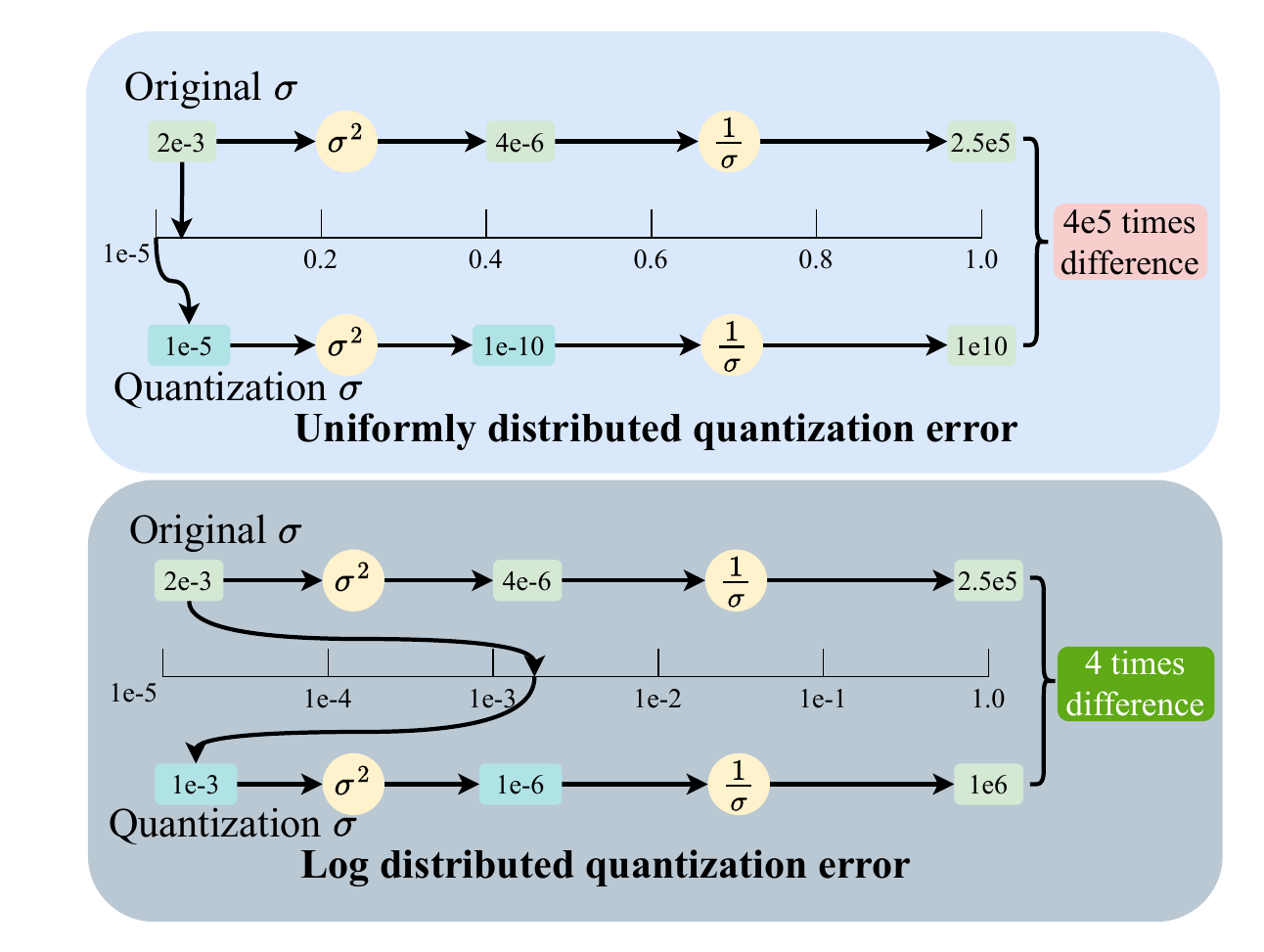}
	\caption{Comparison of quantization error between linear (uniform) quantization and logarithmic quantization of the standard deviation parameter at different scales when running backward propagation. 
    Due to the involvement of $1/\sigma^2$ term in log-probability calculations, quantization errors are significantly amplified for small values. 
    Linear quantization shows substantially higher relative errors in this scenario.}
	\label{fig:challenge:log_quant}
\end{figure}

When switching from simple \textit{SPQ} to include the variational parameters with \textit{VPQ} and \textit{JQ},
further challenges become apparent.
While uniform quantization works well for parameters mirroring the point estimates of deterministic NNs, such as mean for a Gaussian distribution,
the same is not true for more complex parameters, such as the standard deviation of a Gaussian.
In probabilistic programming languages, these parameters are represented in log-space to avoid numerical instabilities.
As a result, uniform quantization can discard substantial information when the parameter values are small.
Figure \ref{fig:challenge:log_quant} illustrates this issue visually. 
We thus opted for a custom logarithmic quantization for these parameters, as highlighted in Fig. \ref{fig:meth:quant}.

Additionally, we find that clipping the weights to a pre-defined range (we recommend $-1$ to $1$) can significantly help in producing high quality uncertainty estimates.
This is in particular noticeable, when a model contains no bias in their linear or convolutional layers.
Here, it leads to improved in-domain classification, as well as better uncertainty calibration.
The optimization has particular importance for quantized BNNs, as the quantization of the bias is in some cases tricky.

All in all, we identify three key points to watch out for when quantizing BNNs:
The activation function must be well-chosen, we recommend SoftPlus for classification tasks.
Parameters that are typically handled in log-space—such as the standard deviation of a Gaussian—should be quantized in log-space as well.
For BNNs without bias terms, the weights should be clipped to a fixed range.

\section{Experimental results}\label{sec:exp}
So far we have set up a comprehensive framework for the quantization of BNNs (Section \ref{sec:multi-level}),
with optimizations for practical deployments (Section \ref{sec:challenge}).
We will now investigate how a BNN reacts to increasingly aggressive bitwidths.
The results shown here are for the Joint Quantization of both variational parameters and sampling results, as this is the most complex and difficult case.
We expand on the result of \textit{VPQ} and \textit{SPQ} further below.

\begin{figure*}
    \centering
	\subfloat[2-bit]{
        \includegraphics[width=0.24\textwidth]{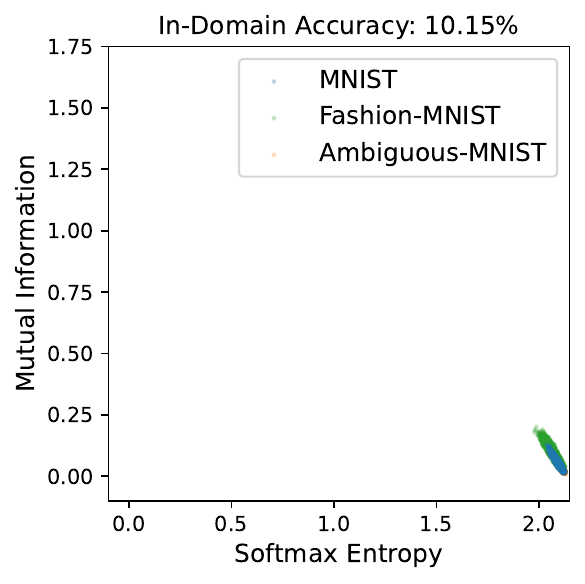}
        \label{fig:exp:2bit}
    }
    \subfloat[3-bit]{
        \includegraphics[width=0.24\textwidth]{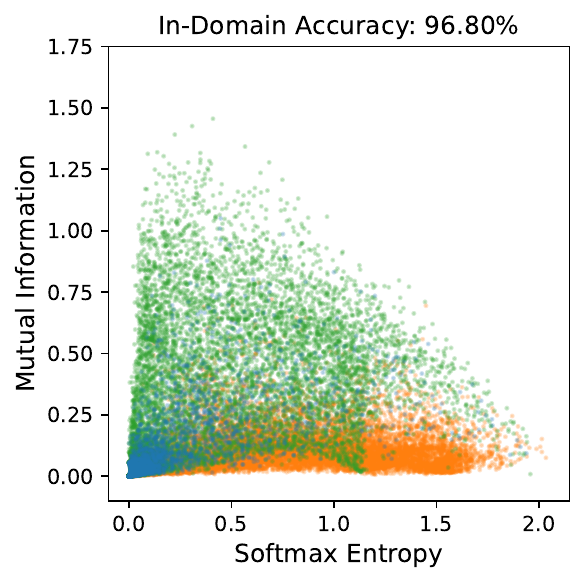}
        \label{fig:exp:3bit}
    }
    \subfloat[4-bit]{
        \includegraphics[width=0.24\textwidth]{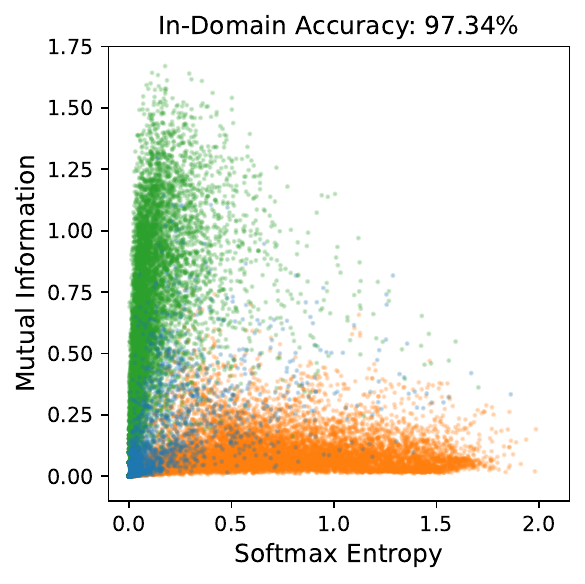}
        \label{fig:exp:4bit}
    }
	\subfloat[32-bit (floating point)]{
        \includegraphics[width=0.24\textwidth]{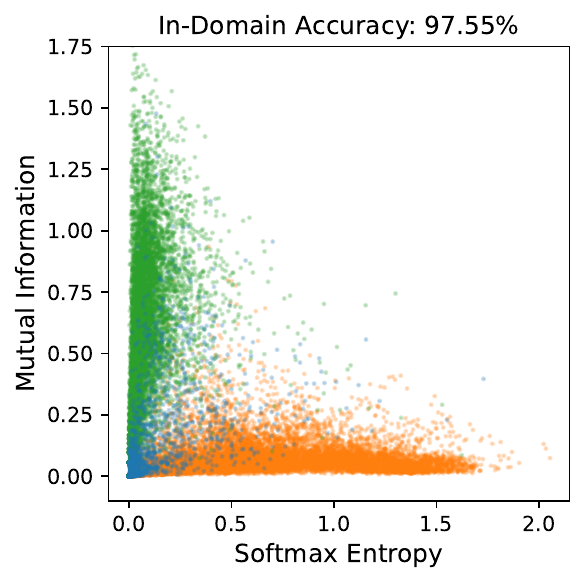}
        \label{fig:exp:float}
    }
	\caption{Scatter plots of Softmax Entropy versus Mutual Information across multiple MNIST variants using the Joint Quantization method.
	Each subplot corresponds to a different bit-width configuration, ordered from left to right such that quantization artifacts are strongest on the left and full 32-bit precision is shown on the right.
	While both in-domain accuracy and uncertainty disentanglement degrade under more aggressive quantization,
	uncertainty disentanglement already deteriorates substantially at 3-bit precision.}
    \label{fig:exp:full_compare}
\end{figure*}

The evaluation is done using the Dirty-MNIST benchmark. 
We train an MLP with two hidden layers of 100 neurons each and SoftPlus activation functions.
The model is first pretrained as a deterministic floating point model for 1500 epochs and learning rate of $10^{-3}$.
The resulting parameters are then transferred into the quantized BNN model of same structure, as a starting point for the Gaussian means ($\mu$).
Finally, the quantized BNN is then trained for 1000 epochs.
We additionally regularize the loss using KL-annealing using a linearly increasing schedule from $0$ to $0.25$ over the BNN training, as described in \cite{klein2025phd}.
Final evaluation is done with 100 samples per test image.

Figure \ref{fig:exp:full_compare} highlights four different quantization bit-widths.
Starting with the floating point baseline in Fig. \ref{fig:exp:float}, one can observe that the BNN starts out with very strong capability to distinguish ID and OOD data, as well being able to successfully disentangle aleatoric and epistemic uncertainty,
while exhibiting good in-domain accuracy.
This performance is reasonably well maintained down to 4 bits, as can be seen in Fig. \ref{fig:exp:4bit}.
At 3 bits the BNNs performance starts deteriorating significantly.
While the in-domain accuracy is only marginally affected, this is not the case for the uncertainty estimation.
In general one can still observe high uncertainties for both Fashion-MNIST and Ambiguous-MNIST,
however the two types of uncertainties are now partially entangled with each other and no clear separation is possible anymore.
At 2 bit quantization the BNN breaks down completely.
Predictions collapse to a very narrow space of MI and SE and in-domain accuracy deteriorates to random guessing.
These results are mirrored in both the Accuracy and AUROC results presented in Table \ref{tab:quantization_results}.
\begin{table}[]
    \centering
    \caption{Quantization performance across different bit-widths and quantization methods. Crossed out cells indicate, that prediction performance had collapsed.}
    \label{tab:quantization_results}
    \begin{tabular}{|lcccc|}
    \hline
    \multicolumn{1}{|l|}{\textbf{Metric [\%]}}                 & \textbf{2-bits} & \textbf{3-bits} & \textbf{4-bits} & \textbf{Full-precision} \\ \hline
        \multicolumn{5}{|c|}{\textbf{Joint Quantization (JQ)}}                                                                          \\ \hline
        \multicolumn{1}{|l|}{Accuracy}                       & \cross{.5cm}{10.15}          & 96.80          & 97.34          & 97.55                  \\
        \multicolumn{1}{|l|}{AUROC: F-MNIST}              & \cross{.5cm}{93.34}          & 74.66          & 86.41          & 84.70                  \\
        \multicolumn{1}{|l|}{AUROC: A-MNIST} & \cross{.5cm}{78.10}          & 94.94          & 96.16          & 97.01                  \\ \hline
        \multicolumn{5}{|c|}{\textbf{Variational Parameter Quantizaiton (VPQ)}}                                                                                  \\ \hline
        \multicolumn{1}{|l|}{Accuracy}                       & \cross{.5cm}{10.61}          & 96.51          & 97.39          & 97.55                  \\
        \multicolumn{1}{|l|}{AUROC: F-MNIST}              & \cross{.5cm}{62.77}          & 93.12          & 84.55          & 84.70                  \\
        \multicolumn{1}{|l|}{AUROC: A-MNIST} & \cross{.5cm}{53.34}          & 94.14          & 96.50          & 97.01                  \\ \hline
        \multicolumn{5}{|c|}{\textbf{Sample Quantization (SQ)}}                                                                                   \\ \hline
        \multicolumn{1}{|l|}{Accuracy}                       & 96.35          & 97.26          & 97.66          & 97.55                  \\
        \multicolumn{1}{|l|}{AUROC: F-MNIST}              & 59.91          & 77.93          & 75.36          & 84.70                  \\
        \multicolumn{1}{|l|}{AUROC: A-MNIST} & 94.70          & 95.91          & 96.20          & 97.01                  \\ \hline
        
    \end{tabular}
\end{table}

For \textit{VPQ} these observations are generally the same, as shown in table \ref{tab:quantization_results}. 
Both ID-Accuracy and uncertainty disentanglement degrade from four to three bit and collapse at two bit.
\textit{SQ} however is more resistant. 
Here the degradation of both metrics continues down to two bits, where uncertainties mix similarly to \ref{fig:exp:3bit}, without fully collapsing.
We attribute this increased resistance to the fact, that the actual variational parameters are still represented at high-precision and are thus able to approximate a complex posterior distribution.

While these experiments were done only on Dirty-MNIST, they show a general trend and guideline, for initial experiments on more complex datasets.

\section{Summary and Outlook}

We proposed a multi-level quantization framework for SVI--based Bayesian Neural Networks and instantiated it with Variational Parameter Quantization (VPQ), Sampled Parameter Quantization (SPQ), and Joint Quantization (JQ). 

This separation of quantization locations shows that BNNs can tolerate aggressive precision reduction while maintaining useful uncertainty estimates: 4-bit joint quantization largely preserves accuracy and the separation of aleatoric and epistemic uncertainty on Dirty-MNIST, whereas at 3 bits this separation deteriorates and at 2 bits both accuracy and uncertainty collapse. 
These findings complement prior work on Post-Training Quantization of Bayesian models by revealing that the location and structure of quantization within the SVI pipeline are as important as the nominal bit-width.
As such this work, as a first, enables multi-level quantization using Quantization-Aware-Training.

Our results yield compact design rules for low-precision BNNs: (i) smooth activations such as SoftPlus improve uncertainty disentanglement compared to ReLU under quantization; (ii) parameters represented in log-space, in particular standard deviations, should use logarithmic rather than uniform quantizers; and (iii) simple magnitude clipping stabilizes training and improves calibration in bias-free architectures. 
These choices together enable uncertainty-aware BNNs at 4-bit precision.

From a systems perspective, the observed robustness suggests that SVI-based BNNs can be compressed by roughly an order of magnitude in parameter memory with modest loss in quality, making Bayesian inference more attractive for edge and accelerator-based platforms, including analog Bayesian machines~\cite{chaotic_nature_2024,photonic_prob_ai_outlook_2025,TWI_pre_print_2025} that naturally operate at low precision and exploit stochastic primitives.

Future work includes extending the study to convolutional and transformer-based BNNs, integrating activation quantization and full-integer inference, and validating the approach on FPGA or analog prototypes, ideally coupled with automated, hardware-aware bit-width allocation, forming a promising path toward truly resource-efficient, uncertainty-aware learning.

\section*{Acknowledgment}
The authors acknowledge support by the state of Baden-Württemberg through bwHPC
and the German Research Foundation (DFG) through grant INST 35/1597-1 FUGG.

\bibliographystyle{IEEEtran}
\bibliography{main}

\end{document}